# IMAGE ACQUISITION IN AN UNDERWATER VISION SYSTEM WITH NIR AND VIS ILLUMINATION

Wojciech Biegański and Andrzej Kasiński

Institute of Control and Information Engineering,
Poznań University of Technology, Poznań, Poland
wojciech.bieganski@doctorate.put.poznan.pl

## ABSTRACT

*The paper describes the image acquisition system able to capture images in two separated bands of light, used to underwater autonomous navigation. The channels are: the visible light spectrum and near infrared spectrum. The characteristics of natural, underwater environment were also described together with the process of the underwater image creation. The results of an experiment with comparison of selected images acquired in these channels are discussed.*

## 1. INTRODUCTION

An autonomous underwater vehicle navigation could be supported by using sonar systems, dead reckoning systems and by using the computer vision [9]. The paper focuses on visual navigation, especially on improving the quality of 2D images of underwater objects. The proposed system captures images in two channels of the light spectrum. The basic assumption is that the images recorded in each band of the wavelength consist of different image features or areas. The channels are the optical spectrum (visual spectrum of light, VIS) and the near infrared band (NIR). The operational environment of designed system are inland waters (lakes or rivers), both natural and artificial, where the visibility in extremal cases, in some areas reaches no more than 20 cm (in lowland rivers). The system is designed to reduce the impact of the infavourable effects influencing the underwater image formation. The acquired images are next combined together resulting in an image consisting of more useful information than any of the two component images. The operation is called the *single-sensor image fusion*.

The hardware used for the tests presented in this paper is a trinocular vision system (TVS) designed and built for the use of inland, underwater imaging [2]. The TVS acquires images in three channels of the light spectrum: NIR, VIS and NUV (near ultraviolet). The comparison of selected images captured in the NIR and VIS channels is presented in this article.

Apart from being a sensory system for the navigation of AUVs, the exact purpose of the designed vision system depends on the kind of mission to execute. In remote mode (with the participation of the operator) the TVS could be used to support searching and rescue missions, inspection of underwater constructions and cataloguing of plants or underwater creatures.

The described vision system is a part of The Isfar Project - a hybrid of an AUV and mini-ROV class vehicle for the exploration of the inland waters [19].

## 2. WATER OPTICS

Natural water is an environment difficult to describe due to its various composition that is not fully identified. The nature of the underwater optical effects is strongly selective and volatile. The intensity of those phenomena has a spatial character, it depends on the location within the the water body (the depth and also horizontal position). Furthermore, the intensity of the effects concerning underwater optics has a temporal character i.e. it could change during the day/night cycle and also it is seasonal. The optical water properties may also change within several years [10]. The constituents found in waters and wastewaters are divided into categories [1]:

- strongly absorbs light, particularly blue, scattering is negligible,
- total suspensoids - responsible for almost all scattering,
- mineral suspensoids - scatter light intensely but usually absorb light weakly,
- detritus - spectral absorption similar to yellow substance, also scatters light,
- phytoplankton - absorbs the light strongly with spectral selectivity and also scatters the light strongly.

The two most significant effects influencing the optical parameters of underwater environment are: light absorption and light scattering, both of them rely on the composition of the underwater environment.

### 2.1. Light absorption

When a photon hits a water molecule it makes that molecule oscillate, hence changes its energy level. The photon is being absorbed during the change of the energy level of the molecule. As a consequence, the radiance of the emitted light drops logarithmically as the distance from the light source grows (Lambert's Law).

The light absorption effect is described by the light absorption coefficient $a$. The intensity of the absorpbtion effect strongly depends on the kind of molecules found in the optical path. The absorption coefficient $a$ grows towards the light of lower wavelengths (IR). Absorption of light by water is minimal within the $\lambda$=400 to 500 nm (violet to green) range [8].

### 2.2. Light scattering

The second effect influencing the transmission of the light by water is the scaterring effect. The effect occurs when the light beam changes its direction while come across the area of the non-water substance found in the optical path. The photons are being re-radiated in any or all directions with unchanged (molecular scattering) or lower (fluorescence) energy content. The last type of scattering is connected with light diffraction, refraction or reflection from suspended particles [1].

The scattering effect is the dominating effect especially in natural waters, because of the diversity and volume of constituents suspended (SOM, suspended organic matter) or dissolved (DOM, dissolved organic matter) in the environment. Moreover in natural waters the light scattering is isotropic i.e. light is scattared in every direction, even towards the light source (known as *backscattering*).

The scattering is described by the scattering coefficient $b$ and the volume scattering function $\beta(\theta)$ (describing the intensity or radiance of light being scattered into the direction of the $\theta$ angle). The effect has significant impact on the transmission of the light by water especially for the light of shorter wavelength.

## 2.3. Light attenuation

The scattering and the absorption effects are inseparable in natural waters. The *a* and *b* coefficients are very difficult to measure apart. The absorption coefficient could only be measured without the influence of the scattering error in very clean water [3].

The combination of the scattering and absorption effects result in the attenuation of the light in underwater environment. The light attenuation is described by the beam attenuation coefficient *c=a+b*, it is the one of the fundamental parameters of the water quality describing its clarity.

The optical parameters *a*, *b*, *c* and *β(θ)* are so-called *inherent optical properties* that fully specify the optical character of the water.

The most tangible and easy to measure parameter describing the optics of any light-transmitting environment is the optical transmission (or transmittance). Transmission is a ratio between the radiance of light emited by the light source ($L_0$) and the radiance of the light measured at the distance *r* from the light source ($L_r$) expressed in percent:

$$T = \frac{L_r}{L_0} \cdot 100$$

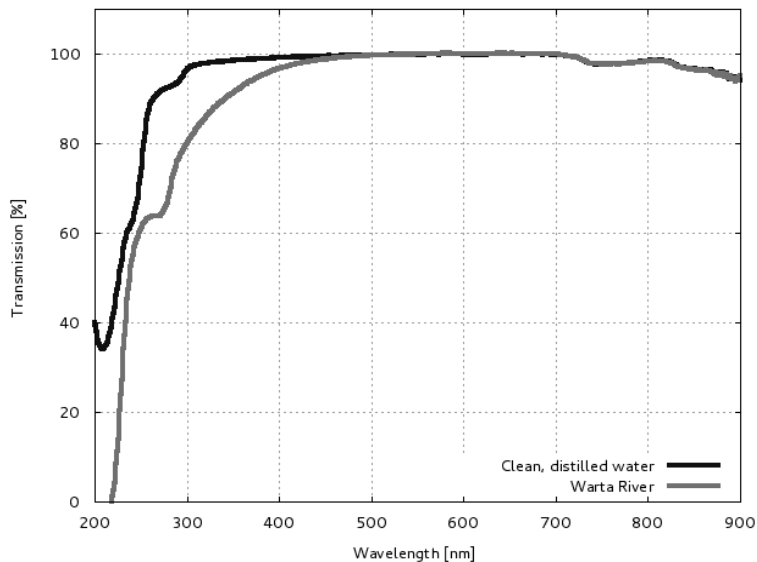

Figure 1. Optical transmission of distilled water nad water taken from Warta River near Poznań, Poland

The transmission of water samples containing the distilled water and the water taken from the Warta river, measured with the use of the spectrophotometer is presented in Fig. 1. The main differences on the graph could be observed in the 200 to 400 nm range of the light spectrum. The differences result from the attendance of the scattering effect in natural water. In the visual range (the wavelength of $\lambda$=400 to 700 nm) the transmission was invariable, still better for the distilled water. The presence of the light absorption effect occured over the $\lambda$=700 nm (red to infrared), for both samples in an equal degree.

## 2.3. Underwater image formation

The optical path of photons emitted from the light source (Power-LEDs), through the object of interest immersed in the underwater environment, to the detector (a CCD camera) is shown in the diagram in Fig. 2. Other optical effects taking part in underwater imaging are: the reflection of the light rays on the surface of the glass viewfinder (two times, from both sides of the viewfinder), the refraction between air/glass and glass/water interfaces (also two times), the reflection on the surface of the detected object, letting actually *see* that object and the distortion of the lens and optical filters.

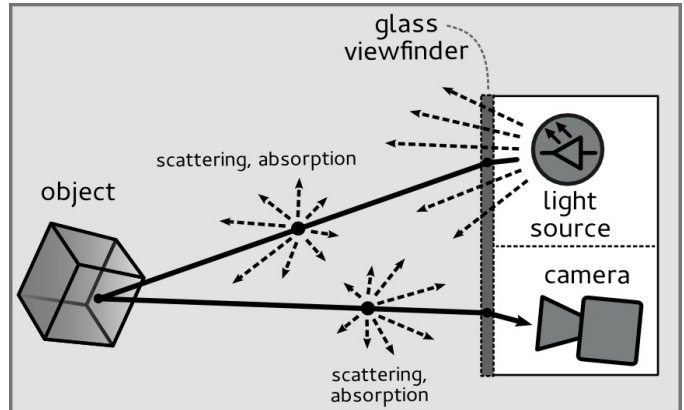

Figure 2. A process of image formation underwater

## 3. UNDERWATER IMAGE ENHANCEMENTS METHODS

The methods of underwater image enhanement (also called image denoising or dehazing) could be divided into two categories: methods that uses software pre-processing algorithms and hardware methods basing on the modifications of the parameters of the optical path of the light.

### 3.1. Software methods

The software methods of the underwater image enhanecement depend mainly on image filtering: homomorphic filtering, anisotropic filtering or filtering in the wavelet domain [14], [18]. Other methods include image deconvolution, contrast equalization [14],[15] and local histogram equalization [13].

### 3.2. Physical methods

Hardware methods of the underwater image enhancement base on the modifications of the parameters of the optical path e.g. by using polarizers [17]. Some experiments on the the various placement of the light sources or with multi-directional fusion were also conducted [16].

A method which initial experiments were presented in this paper connects both hardware and software image enhancement methods.

## 4. CHANNELS OF IMAGE ACQUISITION

The image is captured in two separated channels of the light spectrum. The separation of channels is assured by using the optical filters. The light sources were selected accordingly to the desired wavebands. The system consist of one camera, three optical filters and Power-LEDs as a lighting source. The optical power of the light sources was approximately levelled, since the Power-LEDs of the same electrical power differs in optical power depending on the wavalength of the emitted light, the NIR LEDs have weaker optical power than white LEDs. The filters were mounted on a rotating disk driven by a servomechanism in front of the camera lens, letting to switch of the image acquisition channel. The viewfinder was made of BK-7 (borosilicate) tempered glass.

### 4.1. NIR channel

The near infrared spectrum is a wavelength between $\lambda=750$ nm and $\lambda=1400$ nm. There was a Schott RG-712 long pass filter used (the filter cuts all wavelengths below $\lambda=712$ nm and the Edixeon EDEI-1FA3 Power-LED (maximum optical power at $\lambda=850$ nm).

The optical parameters of natural water in NIR range differs from the parameters occuring in the VIS spectrum [12]. The light in the NIR range of the light is almost impervious to the influence of the scattering effects [11]. On the other hand the light in NIR spectrum is strongly affected by the light absorbtion in underwater environment [5]. The intensity of the absorption effect could depend not only on the molecular structure of the water (and dissolved/suspended substances) but also on the temperature [6], [7].

Another application of the NIR radiation was to use NIR light emitters together with the camera to observe fish. The NIR light is invisible to the them, thus the observation system does not have a notable impact on fish behaviour [4].

### 4.2. VIS channel

The optical (or visual) spectrum of the light were $\lambda$ is situated between 380 and 780 nm. There is a pair of optical filters used in this channel: UVK-2510 UV cut-off filter and ICF-2510 IR cut off filter. The illumination comes from 3-Watt Power-LED emitting the warm white light.

### 5. EXPERIMENT AND RESULTS

The experiment consisted of acquiring a sequence of images in both NIR and VIS channels of the immersed object in order to compare ans describe differences occuring on the images.

The experiment was conducted in laboratory environment. The water tank was the aquarium with blinded panels. The volume of the aquarium is 250 litres. The submerged object of observation is a 11x11 cm cube, where its every face is made of (or covered by) different material, that could appear in lake or river beds. Those faces are:

- a face with a marker (a chessboard) attatched (used as reference),
- a metal sheet covered with rust,
- a tinplate face,
- a rubber face,
- two fabric-covered faces: with straight lines pattern and with circular blobs.

Apart from the object of interest, there were some underwater plants in the tank, black and white gravel and stones. The water comes from the water supply network. The background was a black PVC sheet.

Since the IR radiation coming from the natural light source is completely absorbed a few cm below the water surface, the authors decided to use artificial lighting only.

Two descriptions of the acquired images were proposed due to image comparison: histograms (intensity analysis) and detected edges (feature diversity analysis). Captured images are presented in Fig. 3a and Fig. 3b. On the first pair of selected images, there are three faces of the cube visible: the reference face, a tinplate face and a metal sheet face covered with rust.

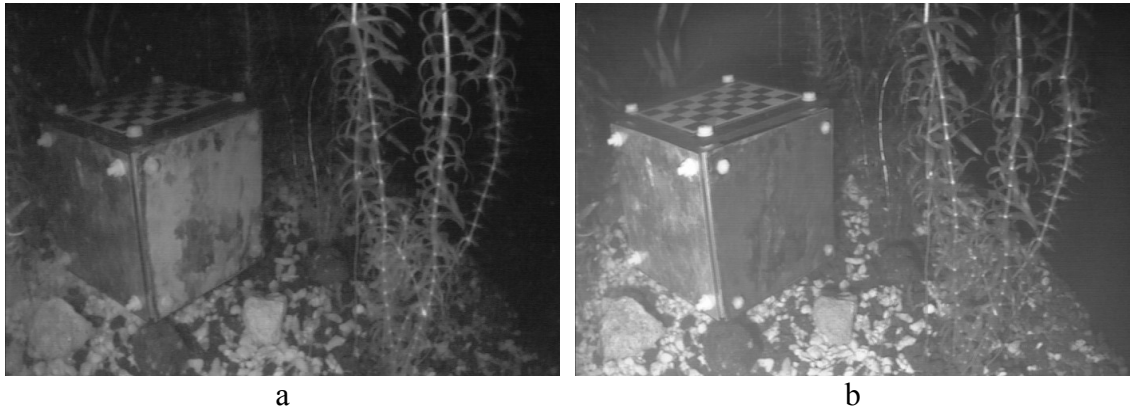

                    a                                                                     b

Figure 3.  Acquired images, a - NIR channel image, b - VIS channel image

The absorption effect causes the images to be less detailed (or darker) depeding on the distance between the source (through object) and the detector. The absorption effect has stronger impact as the wavelength of the emitted light grows, thus the images acquired in the NIR had less brightness than the images in VIS channel.

On the other hand, the scattering effect lowers the contrast of captured images by brightening the water surrounding the space between the object and the detector. Since the scattering effect has stronger impact on the light of shorter wavelengths, it is noticeable on images acquired in VIS channel. Differences mentioned above are visible in the images itself, but also in the histograms presented in Fig. 4.

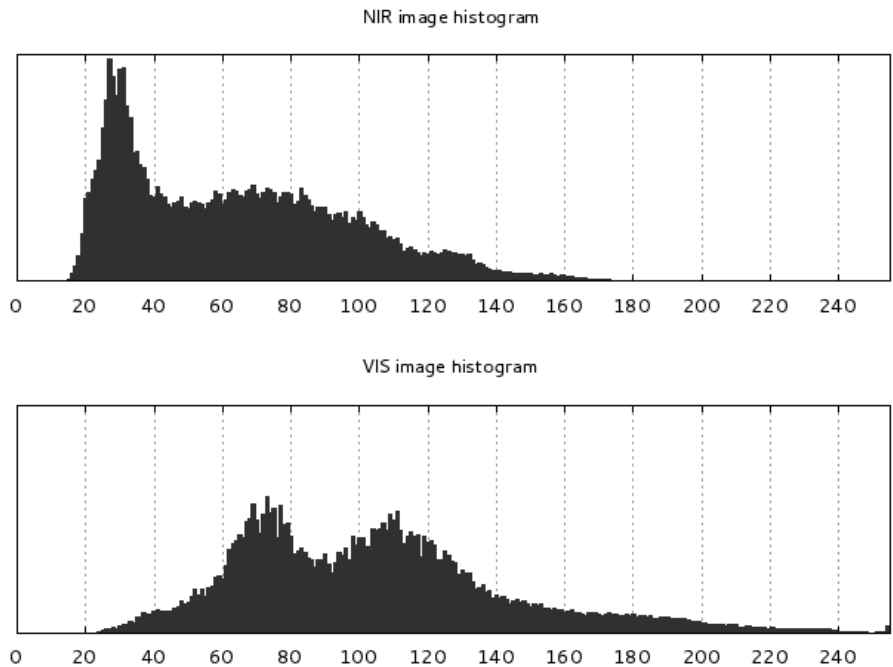

Figure 4.  Histograms of 3a and 3b images

A Canny edge detector with the same threshold values for both images was used. Some edges faded on the NIR images due to the absorption effect (noticeable on the chessboard). The main differences revealed on the area where there were underwater plants. The contrast of plants was better in the NIR channel, hence more edges were detected in NIR (underwater plants reflect the

NIR light, since the energy of IR radiation is not gathered by those plants for the use in the photosynthesis process).

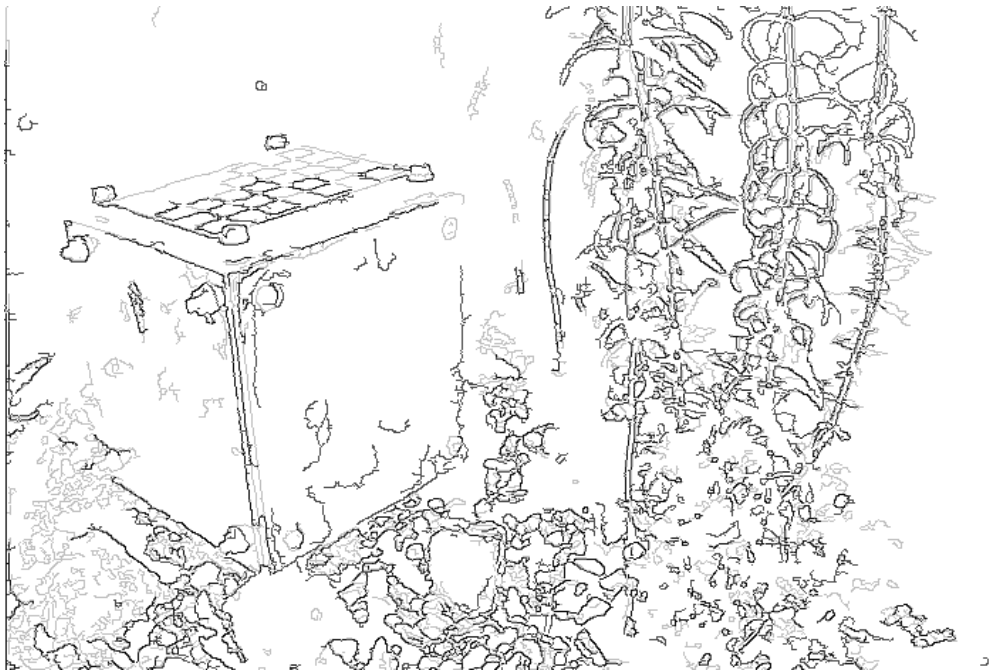
Figure 5. Detected edges, black edges - NIR channel, gray edges – VIS channel

On the second pair of images, there is a cube with three faces visible: a reference face with a chessboard, the face with a metal sheet covered with rust and a face covered with fabric with the circular pattern on it. Images and their histograms are presented in Fig. 5 and Fig. 6.

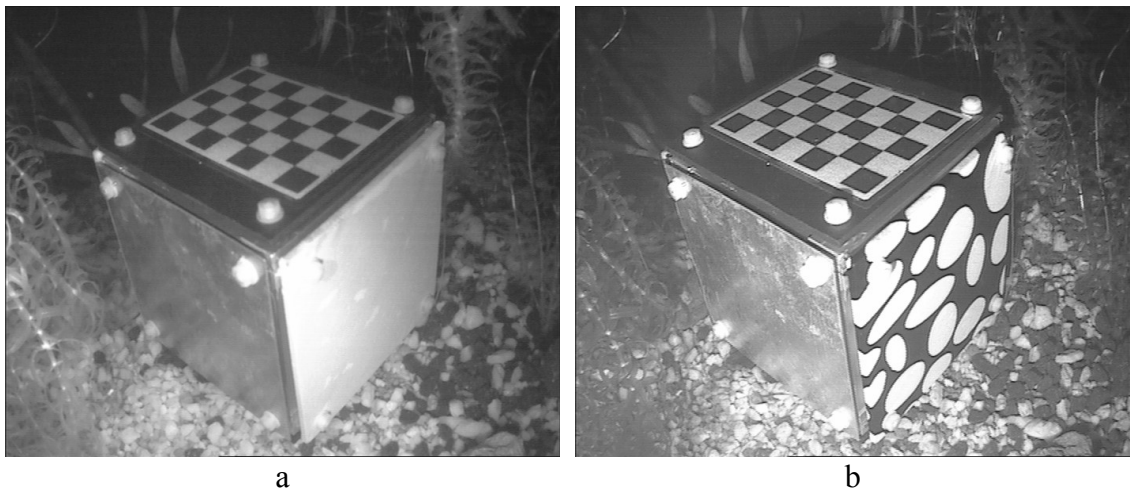
a　　　　　　　　　　　　　　　　b
Figure 5. Acquired images, second pair, a - NIR channel image, b - VIS channel image

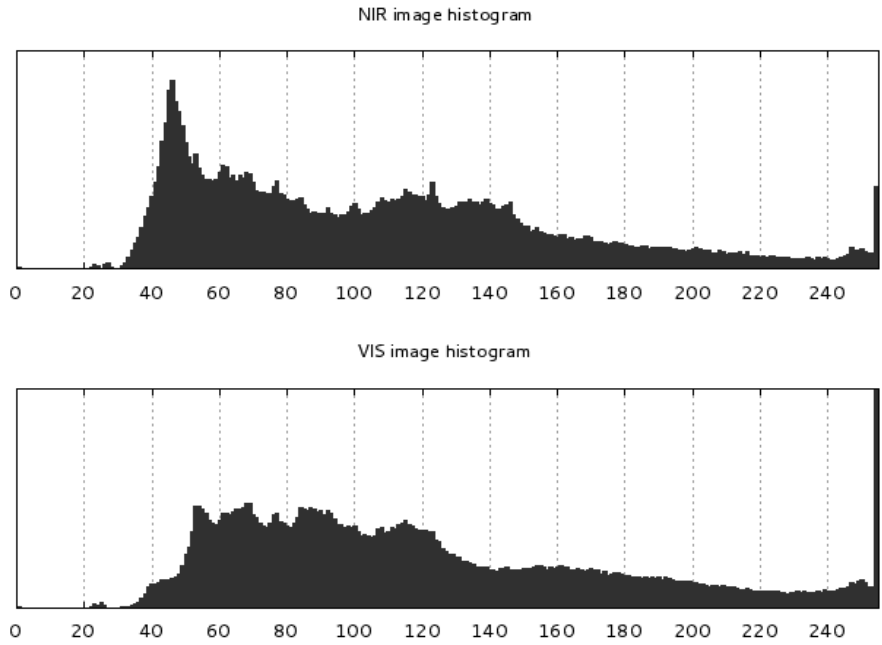

Figure 6.  Histograms of 5a and 5b images

All tested fabrics, no matter what pattern were covered by look similarly in the NIR channel. The circles *seen* on the face of the cube in VIS channel are invisible in NIR channel (NIR light is not affected by the dye used to produce the fabric). The result of the Canny edge detector on the images is shown in Fig. 7. The black fabric, almost invisible in the VIS channel, is detectable in NIR channel, hence the NIR radiation could be used during rescue missions to detect e.g. fragments of clothing.

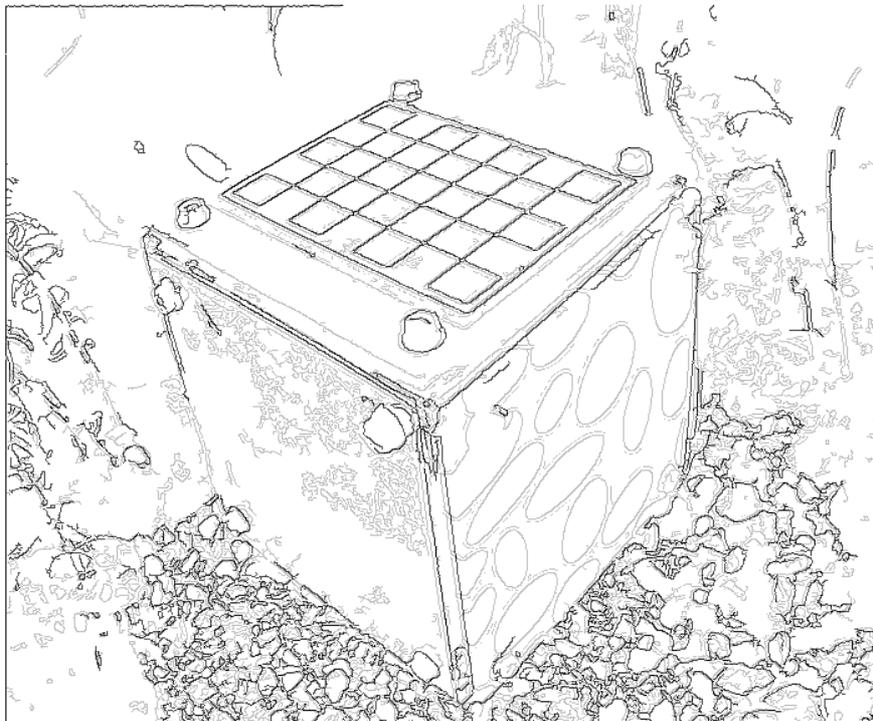

Figure 7.  Detected edges, black edges - NIR channel, gray edges – VIS channel

## 5. CONCLUSIONS AND FUTURE PLANS

The initial tests concerning acquiring of some underwater images with the use of the two channel underwater imaging system were presented in this article. The images were next analyzed in order to find differences resulting from the acquisition channel use. Results obtained in the tests confirmed the presumptions about the principles of radiative transfer in underwater environment.

The underwater imaging with the use of the NIR radiation could find application especially in highly turbid environments such as natural, inland waters due to its resistance to the scattering effect.

Images acquired in NIR include less or equal information than images captured in VIS while imaging objects made of such materials as plastics, metals (some differences in rust-covered surfaces), rubber and, in particular, patterns on fabrics. On the other hand the underwater plants had higher contrast, thus were more distinguishable from the background, than plants captured in VIS channel. A conclusion could be drawn, that imaging with the use of NIR radiation could find be used for searching or cataloguing specific plants. Furthermore, if the presence of plants is undesireable on the images there is a possibility to use the information included in NIR image to remove the plant-filled areas from the VIS image during the image fusion process.

A weighted image fusion algorithm for the images captured in both NIR and VIS channels will be developed, where selected areas on NIR image with assigned weights could be added to (or subtracted from) the VIS image resulting with the image that would be more useful for the navigation algorithms of the underwater vehicle.

Although results are promising, some parts of the system need to be improved. In order to conduct the research concerning the development of the image fusion algorithm, the exact pixel correspondence is needed between both images. In the current system the images were not precisely matched, because the images were not captured simultaneously, since there was a delay needed for switching channels (about 150 ms). The faster servomchanism is required. Moreover, even if the images would be acquired in the same time, there is a pixel disparity resulting from the fact, that camera intrinsics and distortion coefficients are different depending on the channel (due to various refraction coefficients depending on the kind of the filter used). The light sources should need more accurate levelling of the optical power confirmed by prior tests with the use of the radiometer or pyranometer, so that the images would be captured in the same power conditions. Those inconveniences is planned to be removed in short future.

**Authors**

**Wojciech Biegański, MSc. Eng.**
Graduated from the Poznań University of Technology (2009). He is a Ph.D. student at the Institute of Control and Information Engineering of the Poznań University of Technology. His interests are the mobile robotics, especially the visual perception of robots.

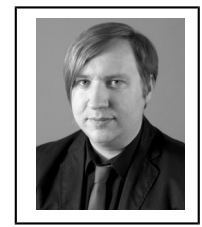

**Andrzej Kasiński, PhD. Eng.**
Graduated from the Poznan University of Technology in 1973 and the Adam Mickiewicz University in 1974. He received the Ph. D. and D. Sc degrees from the Poznan University of Technology in 1979 and 1998, respectively. He was a visiting professor on the Delft University of Technology and the Universidad de Murcia, ENSII Cartagena. Prof. Kasiński has been the head of the Institute of Control and Information Engineering of the Poznan University of Technology since 2002. He is an author of over 150 papers and co-author of 5 patents in the fields of control theory, Pulse-Coupled Neural Network (PCNN), computer vision and biocybernetics.

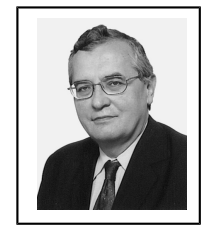